# Building Korean Linguistic Resource for NLU Data Generation of Banking App CS Dialog System

**Jeongwoo Yoon**
DICORA, Hankuk University of Foreign Studies
skyjw1211@gmail.com

**Onyu Park**
DICORA, Hankuk University of Foreign Studies
onyubellapark@gmail.com

**Changhoe Hwang**
DICORA, Hankuk University of Foreign Studies
hch8357@naver.com

**Gwanghoon Yoo**
DICORA, Hankuk University of Foreign Studies
rhkdgns2008@naver.com

**Eric Laporte**
Université Gustave Eiffel
eric.laporte@univ-eiffel.fr

**Jeesun Nam**
DICORA, Hankuk University of Foreign Studies
namjs@hufs.ac.kr

## Abstract

Natural language understanding (NLU) is integral to task-oriented dialog systems, but demands a considerable amount of annotated training data to increase the coverage of diverse utterances. In this study, we report the construction of a linguistic resource named FIAD (Financial Annotated Dataset) and its use to generate a Korean annotated training data for NLU in the banking customer service (CS) domain. By an empirical examination of a corpus of banking app reviews, we identified three linguistic patterns occurring in Korean request utterances: TOPIC (ENTITY, FEATURE), EVENT, and DISCOURSE MARKER. We represented them in LGGs (Local Grammar Graphs) to generate annotated data covering diverse intents and entities. To assess the practicality of the resource, we evaluate the performances of DIET-only (Intent: 0.91 /Topic [entity+feature]: 0.83), DIET+ HANBERT (I:0.94/T:0.85), DIET+ KoBERT (I:0.94/T:0.86), and DIET+ KorBERT (I:0.95/T:0.84) models trained on FIAD-generated data to extract various types of semantic items.

## 1 Introduction

Task-oriented dialogue (TOD) systems are rapidly growing in high demand among various industries seeking ways of improving service quality and coping with customers effectively. TOD systems primarily focus on helping users achieve specific purposes such as hotel accommodation, food order, and product recommendation. Natural Language Understanding (NLU) technology plays a critical role in TOD systems: it helps understanding the information conveyed by user utterances, classifying the intents and filling their slots. For example, in the utterance 카카오 뱅크 계좌 개설해 줘 (*khakhao payngkhu kyeycwa kayselhay cwe*) 'Create a Kakao bank account,' an NLU model has to classify the user's intent as 'request for creating a bank account', but also recognize the argument with the named entity 카카오 뱅크 'Kakao bank', fill it in the 'Bank Entity' slot, and identify 'account' as an entity in banking service.

NLU tasks such as intent classification and slot filling are usually implemented by supervised learning, requiring a large amount of annotated training data. However, it is hard to find publicly available Korean training datasets for TOD oriented NLU, due to privacy issues, and constructing them costs considerable time and human labor. In addition, since users produce diverse utterances with different terminologies and linguistic forms depending on domains, annotated training data should reflect that diversity in each domain. The performance of NLU models is likely to be directly affected by this aspect of the training dataset.

Moreover, as most of the user utterances in TOD are meant to request specific information or actions,

covering the linguistic patterns that occur in directives and questions is essential to increase the NLU coverage of diverse utterances. In Korean, discourse markers used in requests frequently contain certain patterns with word endings and performative predicates, which can be realized in different moods, e.g. -해 줘/봐(*hay cwe/pwa*). In order to classify Korean request utterances, such discourse expressions should be included in the training data.

This study proposes a linguistic resource named Financial Annotated Dataset (FIAD), which allows for generating annotated training data in the banking CS domain. FIAD specifies linguistic forms with extensive linguistic variations, and has been constructed on the basis of a corpus of banking app reviews. It consists of three parts: TOPIC(ENTITY, FEATURE), EVENT, and DISCOURSE-MARKER. Each part allows for generating words and multi-word expressions (MWEs) reflecting the characteristics of the corpus in terms of syntax, semantics, and discourse, and contains a meticulous description of their grammatical constraints in Local Grammar Graphs (LGG) (Gross, 1997, 1999). The combinations of the parts allows a generation of training data adjustable to a required size and linguistic characteristics. FIAD covers various types of discourse markers of directives and questions that are challenging for NLU of Korean sentences.

The TOPIC part is comprised of ENTITY and FEATURE. ENTITY covers named entities, and FEATURE includes common nouns related to services. ENTITY and FEATURE are used to fill slots with detailed information.

EVENT covers utterances expressing intents. It invokes modules of TOPIC to generate utterances in compliance with the syntactic and semantic constraints between an intent and the entities or services mentioned in the same utterance.

DISCOURSE-MARKER contains a variety of discourse expressions with predicate endings and auxiliary verbs. This part is largely domain-agnostic. It is modularized according to the Korean honorific system and sentence moods, and covers direct and indirect speech acts.

A main graph specifies how the expressions described in the three parts can be combined into standalone utterances.

FIAD generates named entities and utterances typical of banking apps or services, along with semantic annotations based on the rich linguistic resources. By selecting some of the LGGs, it is possible to generate NLU data of a given size, or tuned to some given politeness levels. It leads to the advantage of obtaining a vast amount of annotated training data containing typical and grammatically fit utterances with time efficiency and less workforce than by collecting or creating the data through crowdsourcing.

## Related Work

Building a TOD system for a domain without enough available data resources requires collecting a significant amount of training data and carrying out a laborious annotation process. One of the popular open training datasets for TOD is the Airline Travel Information Systems (ATIS) dataset (Hemphill et al., 1990), which consists of utterances about requests for flight information and which is used for automated airline travel inquiry systems.

The Wizard of Oz (WOZ) and template-based methods are popular for building training data for TOD systems. The WOZ method sets 2 participants: the wizard and the user. The wizard pretends to be a dialogue system, and the user does not know the wizard conversing with them is human. Budzianowski et al. (2018) introduces the Multi-Domain Wizard of OZ dataset (MultiWOZ), which contains 10k dialogues and covers fully annotated conversations spanning over 7 domains. Asri et al. (2017) proposes the ‘Frames’ dataset which contains dialogues about booking a trip based on user requirements.

The template-based method sets a fixed template which consists of entity slots and speech acts-related expressions, and generates data by filling entities in slots of the template. Borhanifard et al. (2020) proposes a Persian dialogue dataset for online shopping dialogue generated by the template-based method. A dialogue system trained by using a mixture of template-based generated data and manually annotated data shows a decent performance. Şimşek and Fensel (2018) presents a template-based method for generating training data. The template structure is based on Web API annotation schema collected from ‘schema.org’.

As for Korean data, C. Hwang et al. (2021) proposes a linguistic resource in the financial technology (fintech) domain. The linguistic resource consists of patterns of queries, complaints and requests in that domain, with fine-grained linguistic information, and it allows for generating and annotating question answering data. The Korea Institute of Science and Technology Information releases a Korean conversation dataset on AiHub (2018)[1]. The dataset covers domains involving small businesses and public services, such as restaurant reservation, online shopping, and public transportation.

However, most of the resources, in particular WOZ-based and template-based, are in English. Resources based on the crowdsourcing method initially contain a lot of noise, requiring several data refinements and complex preprocessing, and they tend to have fewer linguistic variations, as they are created by random unprofessional contributors.

Local grammar graphs (LGG) have often been devised to recognize semantic categories of expressions, e.g. time adverbs in Korean (Jung, 2005) or proper names in Arabic (Traboulsi, 2006). A local grammar graph is a directed word graph, or finite-state automaton, with paths labeled by linguistic forms. LGG is a powerful method to describe linguistic patterns with lexical, syntactic, and semantic restrictions in a readable way. A graph may invoke subgraphs, which specify parts of the phrases specified by the graph. In other words, a subgraph fills a slot in a graph. A ‘local grammar’ is usually made up by a collection of graphs.

As opposed to the traditional use of LGGs, which is recognizing phrases in texts, we use them in this research in order to generate linguistic forms. Thus, we implemented a generator of utterances with annotation of slots and intents for a dialogue system. The generator enumerates paths of Local Grammar Graphs (LGGs) and generates large-scale training data covering the Korean honorific system, different grammatical moods, and various speech acts. The generation can be parameterized in order to create training data according to the desired features. Using the training data, we evaluate the performances of DIET models (Bunk et al., 2020) with pre-trained embeddings.

## Methodology

FIAD was constructed in three phases: data analysis, resource construction, and data generation, as illustrated in Figure 1.

Figure 1: FIAD building process

The first phase, *Data Analysis*, is performed by analyzing domain-specific corpus and extracting core keywords that should be recognized. The second phase, *Resource Construction*, consists of building a Deco-Dom and LGGs that contain TOPIC(ENTITY, FEATURE) words, EVENT expressions and DISCOURSE-MARKERS. Finally, the third phase, *Data Generation*, is conducted by the combination of the three modules of the language resources represented in Phase 2.

## 2 Data Analysis

Since collecting users’ dialogue data raises privacy issues, we collected a corpus of banking app reviews as alternative data. 126,598 banking app reviews were collected from Appstore and Playstore. On a scale of 1 to 5, we focused on reviews with a score of 3 or lower because low score reviews tend to include more users’ requests or complaints on banking services than high score reviews.

We used the Mecab-Ko Korean Morphological analyze[2] to split the collected reviews into morphemes. Then, we extracted key morphemes, nouns, predicates, and inflectional endings using their TF-IDF (Term Frequency-Inverse Document Frequency) weight.

Based on the extracted keywords and on the observation of utterance patterns, we set a language resource in three parts: TOPIC(ENTITY, FEATURE), EVENT, and DISCOURSE-MARKER. Each part has submodules which are separated based on their semantic content. All the modules and their sub-modules are organized as shown in Fig. 2. The module/submodule hierarchy means that the expressions specified by the submodule are a subset of those specified by the module. More details on this resource are provided in Section 5.

[1] https://aihub.or.kr/
[2] https://github.com/hephaex/mecab-ko

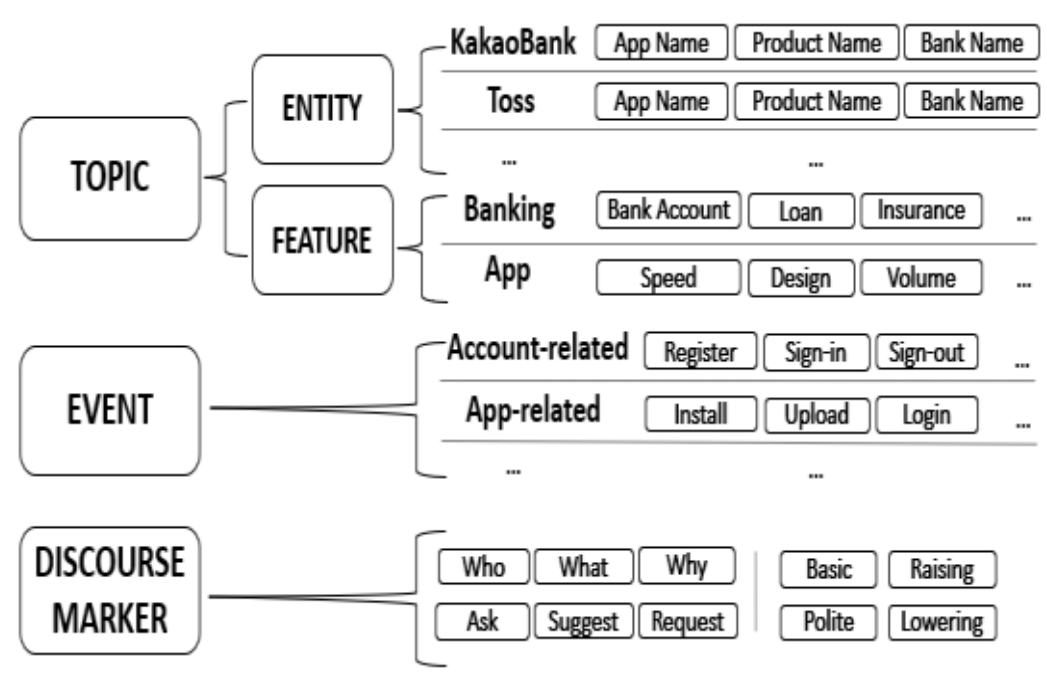

Figure 2: Module and submodule hierarchy

## 3 Resource Construction

### 3.1 TOPIC

We divide the TOPIC part of FIAD into ENTITY and FEATURE for modularity and scalability. ENTITY includes named entities which refer to product names and names of specific banks, and FEATURE includes common nouns frequently used in reviews and dialogues about banking and financial app services. The modules and submodlues of ENTITY and FEATURE are displayed in Tables 1 and 2.

| Category | Entity Submodules | # of patterns |
|---|---|---|
| BankName | KakaoBank, TossBank, etc. | 53 |
| AppName | Kakao Pay, Toss, etc. | 167 |
| ProductName | KakaoBank 26Weeks Deposit, TossBank EmergencyFund, etc. | 1,938 |
| Total | | 2,158 |

Table 1: {ENTITY} modules and submodules

| Category | Feature Submodules | # of patterns |
|---|---|---|
| Banking | bank account, loan, stock, insurance, etc. | 147 |
| App | speed, volume, design, etc. | 281 |
| Total | | 428 |

Table 2: {FEATURE} modules and submodules

The main role of TOPIC is to fill slots. Modules and submodules of ENTITY and FEATURE are used to set slots in the utterances. This allows researchers to change or add training data of a specific domain by selecting ENTITY modules or submodules, such as banks' product names and bank names. For example, *Kakao bank card* can be replaced with *Toss card* by replacing the <KakaoBank> ENTITY submodule with the <Toss> ENTITY submodule.

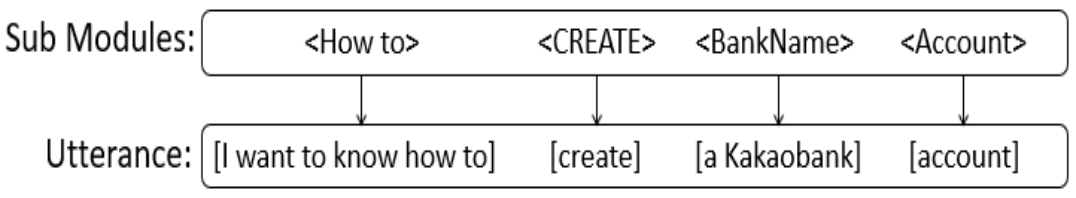

Figure 3: Substituting {ENTITY} Submodules

### 3.2 EVENT

We set EVENT submodules on the basis of key verbs from the review data and semantic restrictions on their arguments. EVENT invokes many TOPIC LGGs, due to semantic restrictions, i.e. because verbs in EVENT require a specific type of TOPIC words or phrases.

(1a) <계좌>를 개설하다 (*kyeycwalul kayselhata*)
CREATE a <bank account>
(1b) *<속도>를 개설하다 *(*soktolul kayselhata*)
*CREATE a <velocity>

(1a) and (1b) show an argument-predicate structure with a noun and a verb. (1a) sounds natural, but (1b) doesn't, because the <velocity> semantic category in (1b) does not combine with the Korean verb used in *CREATE a bank account*. We set specific ENTITY submodules whenever an EVENT verb required it, so as to generate natural utterances.

| Category | EVENT Submodules | # of patterns |
|---|---|---|
| Account | create, sign-in, sign-out, etc. | 510 |
| Banking Product | send, take, put, etc. | 924 |
| Financial Product | buy, sell, management, etc. | 454 |
| App | install, upload, pay, etc. | 942 |
| Total | | 2,830 |

Table 3: {EVENT} modules & submodules

The modules and submodules in Tables 1 to 3 were established on the basis of a semantic analysis of the topics, events and speech acts expressed in the corpus. In this way, 2,158 ENTITY noun phrases, 428 FEATURE noun phrases and 2,830 EVENT predicative patterns are discerned from the banking App reviews introduced in Section 4. They are representative of the contents of the corpus. This work was performed by speakers of the language trained in semantic analysis.

### 3.3 DISCOURSE-MARKER

DISCOURSE-MARKER contains various types of multi-word expressions (MWEs) that represent specific speech acts. MWEs are the expressions that consist of several words and connotate specific meanings that can not be derived from their components: ~해줄 수 있나요? (haycwul swu issnayo) '*Could you ~*' is one of the MWE used when speaker wants to request something to hearer.

#### Request types

In the banking domain, the 'Request' speech act is mainly observed. 'Request' speech acts are further classified in two types: 'Information Request' and 'Action Request'. In this way, the 'Information Request' type is divided into 8 semantic categories, and the 'Action request' type is further divided into 3 categories: 'Dissatisfaction' and 'service error', and 'Demand.' 'Dissatisfaction' and 'Service Error' submodules are not typical types of expressions related to 'Request' speech act, but negative comments to the app and error reportings imply request for fixing bad issues on app functions and services. Expressions of 'Demand' are directly collocated with event phrases, which means 'request for specific action'.

| Request type | Semantic category | Example | # of patterns |
|---|---|---|---|
| Information Request | Person | 누구 (Who) | 2,001,525 |
| | Product | 무엇 (What) | 1,186,482 |
| | Method | 어떻게 (How) | 1,401,355 |
| | Reason | 왜 (Why) | 537,294 |
| | Location | 어디서 (Where) | 1,872,303 |
| | Time | 언제 (When) | 510,081 |
| | Age condition | 몇살 (What age) | 667,860 |
| | Cost/Quantity | 얼마 (How much) | 322,844 |
| Action Request | Dissatisfaction | 짜증 (Annoying) | 560,287 |
| | Service Error | 에러 (Error) | 451,865 |
| | Demand | 희망/요구 (Wish) | 11,482 |
| Total | | | 9,523,378 |

Table 4: 'Request types' in DISCOURSE-MARKER module

#### Sentence types

As an agglutinative language, Korean has a diverse range of inflectional endings with various pragmatic functions: **imperative** endings mainly express imperative 'Request' speech act. However, the 'Request' speech act can also be represented by other sentence types such as **declaratives, interrogatives,** and **suggestives**. Thus, we include four types of discourse endings that represent four sentence types in DISCOURSE-MARKER module as shown in Table 5.

| Sentence type | Examples |
|---|---|
| Declaratives | 계좌 개설 [누가 담당하는지 알고 싶어] ([I want to know who is responsible for] creating an account) 계좌 개설[할래] ([I want to] create a bank account) |
| Imperatives | 계좌 개설 [누가 담당하는지 알려줘] ([Tell me who is responsible for] creating an account) 계좌 개설[해라] (Create a bank account) |
| Interrogatives | 계좌 개설 [누가 담당하는지 알 수 있나]? ([Can I ask you who is responsible for] creating an account) 계좌 개설[할 수 있나]? ([Can you] create a bank account) |
| Suggestives | 계좌 개설 [누가 담당하는지 알아보자] ([Let's figure out who is responsible for] creating a bank account) 계좌 개설[하자] ([Let's] create a bank account) |

Table 5: Sentence types in DISCOURSE-MARKER module

#### Honorific types

Korean has various types of honorific markers. The honorific system reflects the hierarchical and relational organization of Korean society and is marked in verbal endings. Korean speakers form their utterances with the honorific levels appropriate to their relationship with the hearer and with the persons they are mentioning. There are three types of honorific markers in Korean: subject, object, and hearer honorific markers. Among these, the hearer honorific markers play an important role in conversations because they refer to the relationship between speaker and hearer.

Korean hearer honorific markers consist of 6 degrees of speech styles: 합쇼체 '*hapsyo*-style', 하오체 '*hao*-style', 하게체 '*hakey*-style', 해라체 '*hayla*-style', 해요체 '*hayyo*-style', and 해체 '*hay*-style.' The first four speech styles depend on the speaker and hearer's social positions and are used in formal speech. The last two speech styles are used in informal speech when the speaker and hearer enjoy some degree of intimacy (National Institute of Korean Language, 2005) .

In contemporary Korean, two speech levels, 하게체 '*hakey*-style', 하오체 '*hao*-style' are being less used. Therefore, we set four categories for Korean hearer honorifics in the discourse marker module. The examples of four types of honorific markers are in Table 6.

| Formality | Speech Styles | Category | Examples |
|---|---|---|---|
| Formal | 합쇼체 *hapsyo* style | Raising | 계좌개설 [누가 담당합니까]? 'Who is responsible for creating an acccount?' |
| | 하오체 *hao* style | | 계좌개설 [해주십시오] 'Please create an account' |
| | 하게체 *hakey* style | Lowering | 계좌개설 [누가 담당하나]? 'Who is responsible for creating an acccount?' |
| | 해라체 *hayla* style | | 계좌 개설 [해줘라] 'Create an account' |
| Informal | 해요체 *hayyo* style | Polite | 계좌개설 [누가 담당해요]? 'Who is responsible for creating an acccount?' 계좌 개설 [해줘요] 'Please create an account' |
| | 해체 *hay* style | Basic | 계좌개설 [누가 담당해]? 'Who is responsible for creating an acccount?' 계좌 개설 [해줘] 'create an account' |

Table 6: Honorific types in DISCOURSE-MARKER module

The three types represented in Table 4-5-6 are combined together according to lexico-semantic and syntactic restrictions among the elements.

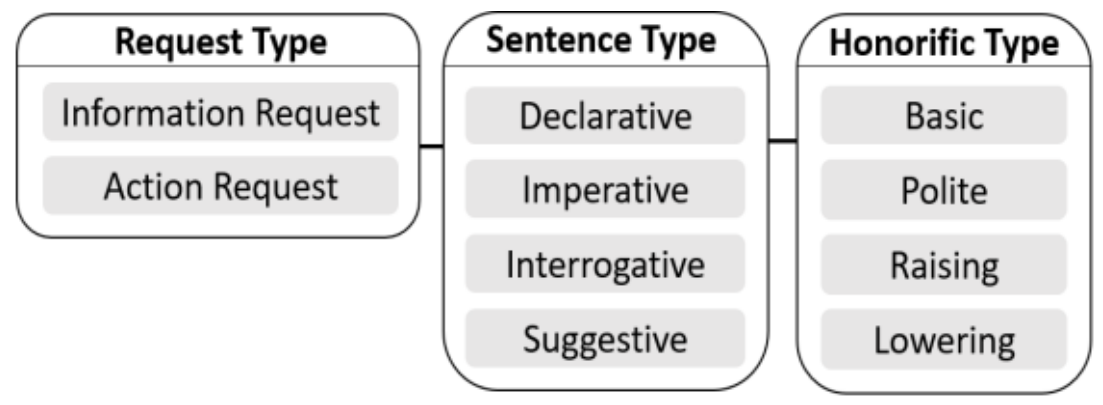

Figure 4: Request/Sentence/Honorific types in DM module

For instance, the combination of '[Information Request]-[Declarative]-[Basic]' generates an example such as '(계좌 개설)하려면 어떻게 해야 하는지 알고싶어. 'I want to know how to do (to create a new account).' and that of '[Action Request]-[Interrogative]-[Raising]' produces an example such as '(계좌 개설)해 주시겠습니까? 'Would you like to (create a new account for me)?'.

## 4 Data Generation

### 4.1 FIAD generated by linking-LGGs

We generated utterance data using linking-LGGs with parameter selection and conjugation postprocessing.

Each module of language resources introduced in Section 5 consists of a set of LGGs, which is compiled into Finite-State Transducers (FST) through the open-source Unitex/GramLab platform (Paumier, 2003). The generator outputs linguistic patterns and their annotations by exploring transitions of the FSTs.

The three parts of the language resources are connected in several ways by the linking-LGG displayed at the top of Figure 5. There are four paths in this graph, which represent such combinations: through this processing, about 60 trillion utterance patterns are generated and registered in FIAD. The submodules of each module are used to annotate the 'TOPIC' and 'EVENT' information.

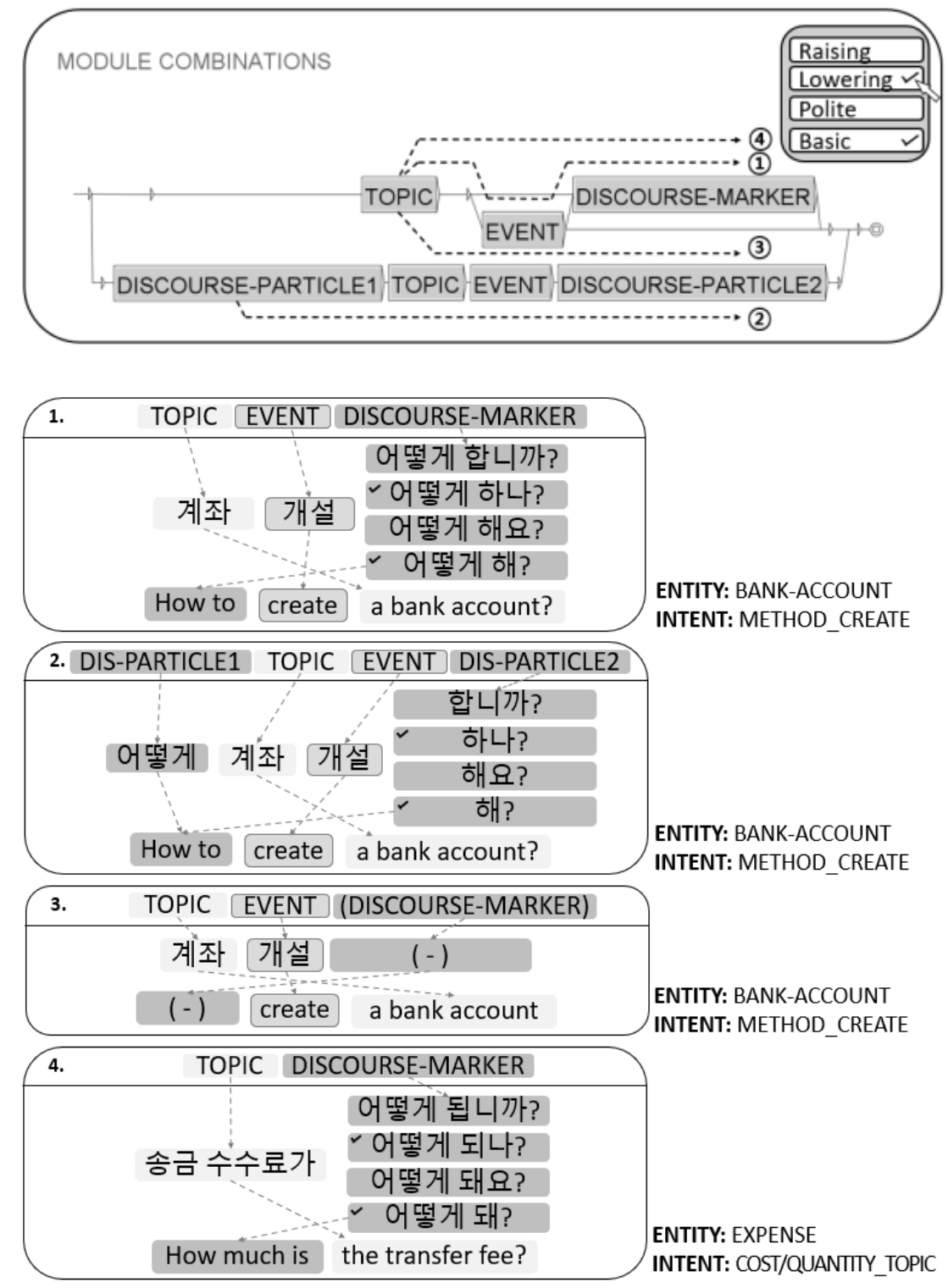

Figure 5: Linking-LGG combining resource modules, and examples of generated results

The first example in Fig. 4 is the basic type of combination, in which TOPIC, EVENT, and DICOURSE-MARKER are connected in sequence.

The second example exhibits a discontinuous discourse marker. As Korean has relatively free word order, parts of discourse expressions, such as 'Wh-words', generated by DISCOURSE-PARTICLE, can occur separately from the rest of the expression.

The third example presents an ellipsis of a dicourse marker. Although DISCOURSE-MARKER is not invoked, the utterance implicitly requests a specific act. Therefore, these utterances are annotated with the 'Request for action.'

The last example shows a case where EVENT is not invoked. The intent of this type of utterance is ‘Request for information.’

### 4.2 Generation of training data

Before the generation of a training data for AI pre-trained language models, we set the speech styles of the utterances by selecting DISCOURSE-MARKER modules depending on the speech styles that the desired TOD system should be able to process. We also set the number of utterances of the training data. We used equation (1) to control the number of utterances. The equation (1) assigns a weight to each expression used in the DISCOURSE-MARKER module, and we generate the utterances with the highest weight to reach the specified size of the training data.

$$w_n = log_2(1 + (syl_{max} - syl_n)) \qquad (1)$$

Equation (1) computes the weight of an utterance, which calculates the priorities between utterances. $syl_{max}$ is the maximum length, in syllables, of the expressions in a DISCOURSE-MARKER module. $syl_n$ is the length of the current discourse marker. The formula applies a logarithmic scale to the difference, for normalization. Using this formula, we gives priority to shorter utterances over longer utterances, and this policy is grounded in economy of language. Everyday experience shows that speakers naturally tend to avoid redundancy and use shorter utterances when possible, although this rule is not absolute.

After speech style and data size selection, we generated training data by recursive exploration of the transitions in each LGG module. Since Korean is an agglutinative language, the generation of linguistic forms requires a description of conjugations of verbs. To apply conjugation rules to the verbs, we used the conjugation class information in Dictionnaire Électronique du Coreen (DECO) (Nam, 2018), a Korean lexical database.

## 5 Experiments

### 5.1 Comparison with DIET performance

To evaluate FIAD, we used the RASA open source framework [3] and FIAD-generated data to build NLU models that detect and classify intents and entities in utterances. RASA provides a Dual Intent Entity Transformers (DIET) classifier and flexible training pipelines.

We experimented several pipelines to test the performance of NLU models. Each pipeline uses the Open Korean Text Tokenizer (Okt).[4] As to the DIET model, we used the KoRASA hyper-parameters for Korean (M. Hwang et al., 2021), listed under ‘DIET-Opt’ in Table 7.

| **Parameter** | **DIET-Base** | **DIET-Opt** |
| --- | --- | --- |
| Epoch | 300 | 500 |
| Transformer layers | 2 | 4 |
| Transformer size | 256 | 256 |
| Connection density | 0.2 | 0.3 |
| Embedding dimension | 20 | 30 |
| Hidden layer size | [256, 128] | [512, 128] |

Table 7: Hyperparameters for the DIET-Base and DIET-Opt models

As to the training data, we selected 107 types of intents highly used in the review data. We used 90,999 sequences generated by FIAD with these types of intents.

As to the test data, native speakers created 1,000 TOD utterances about banking CS and annotated the intent and entity slots for each utterance. To compute the scores, we used a weighted average in accordance with the proportions of the intents and the slots. Since there are no comparable datasets with which we can test our dataset, we set the perfomance of DIET classifier with fine-tuned BERT model, trained by NLU-Benchmark dataset (Liu et al. 2019), as a baseline. The baseline result and performance of DIET model with FIAD are illustrated in Table 8.

| Model | **Tag** | **Precision** | **Recall** | **F1 score** |
| --- | --- | --- | --- | --- |
| DIET+BERT (Baseline) | Intent | 89.67 | 89.67 | 89.67 |
| | Entity | **86.78** | 84.71 | **85.73** |
| DIET (FIAD) | Intent | **0.9278** | **0.9140** | **0.9142** |
| | Entity | 0.8256 | **0.8760** | 0.8377 |

Table 8: Experiment results on FIAD

Although the FIAD is trained without fine-tuned BERT, our model generally outperformed in intent analysis. However, the results showed similar or slightly lower performance in entity analysis: it is due to the TOPIC combinations of ENTITY and

[3] https://rasa.com

[4] https://github.com/twitter/twitter-korean-text

FEATURE that raise the complexity of topic recognition.

### 5.2 Performance of Pretrained Models

In order to check the improvement of the performance, we include pre-trained embeddings in the pipelines. We compared three different Bidirectional Encoder Representations from Transformers (BERT) models for Korean: HanBERT, [5] KoBERT, [6] and KorBERT. [7] These models differ in their vocabulary size and training data. Table 9 shows the result of the evaluation.

| Model | Tag | Precision | Recall | F1 score |
|---|---|---|---|---|
| DIET+HanBERT | Intent | 0.9504 | 0.9440 | 0.9421 |
| | Entity | 0.8465 | 0.8809 | 0.8566 |
| DIET+KoBERT | Intent | **0.9616** | 0.9510 | 0.9493 |
| | Entity | 0.8506 | **0.8952** | **0.8675** |
| DIET+KorBERT | Intent | 0.9613 | **0.9530** | **0.9525** |
| | Entity | **0.8507** | 0.8798 | 0.8486 |

Table 9: Experiment results on RASA NLU

The 'DIET+KoBERT' model shows the highest f1-score in entity extraction, and 'DIET+KorBERT' shows the highest f1-score in intent extraction. Thus, the experiments underline that the pipelines with pre-trained embedding outperform the pipeline without pre-trained embedding. The results show that FIAD-generated training data allow for training models to efficiently extract intents and entities from utterances.

## 6 Conclusion

This study presents a linguistic resource named FIAD and its use to efficiently construct NLU training data for Korean banking CS TOD systems. FIAD consists of LGGs and comprises three parts: TOPIC(ENTITY, FEATURE), EVENT, and DISCOURSE-MARKER. Each part contains modules dedicated to semantic categories, and generates utterances where intents and entity slots are annotated as belonging to these semantic categories.

TOPIC(ENTITY, FEATURE) generates expressions used as 'slots' of intents. EVENT provides utterances with an intent and its arguments realized as slots, taking into account semantic restrictions. DISCOURSE-MARKER includes diverse MWEs which represent speech acts, sorted by moods and speech styles, in compliance with the Korean system of honorific markers. These three parts are combined in four ways into four types of utterance patterns for the training data generation.

The models have a clear benefit from the pre-trained embeddings. The performance of the DIET+KorBERT models trained on FIAD-generated data shows 0.86 and 0.95 f1-score in entity and intent extraction, respectively, showing that the concept is applicable. In addition, the modular structure of FIAD offers flexibility for building extensive training data adapted to specific aims, by changing the order of the components or by selecting the modules to focus on particular types of entities, intents or speech styles.

## Acknowledgments

This work was partially supported by LINITO Ltd. (http://linito.kr) and DICORA Research Center (http://dicora.kr) in Hankuk University of Foreign Studies. We thank the anonymous reviewers for their helpful comments.

## References

Zeinab Borhanifard, Hossein Basafa, Seyedeh Zahra Razavi, and Heshaam Faili. 2020. Persian Language Understanding in Task-Oriented Dialogue System for Online Shopping. In *Proceedings of IKT, the 11th International Conference on Information and Knowledge Technology*, pages 79-84.

Paweł Budzianowski, Tsung-Hsien Wen, Bo-Hsiang Tseng, Iñigo Casanueva, Stefan Ultes, Osman Ramadan, and Milica Gašić. 2018. MultiWOZ - A Large-Scale Multi-Domain Wizard-of-Oz Dataset for Task-Oriented Dialogue Modelling. In *Proceedings of the 2018 Conference on Empirical Methods in Natural Language Processing*, pages 5016–5026, Brussels, Belgium. Association for Computational Linguistics.

Tanja Bunk, Daksh Varshneya, Vladimir Vlasov, and Alan Nichol. 2020. *DIET: Lightweight Language Understanding for Dialogue Systems*. ArXiv, abs/2004.09936.

Layla El Asri, Hannes Schulz, Shikhar Sharma, Jeremie Zumer, Justin Harris, Emery Fine, Rahul

[5] https://github.com/monologg/HanBert-Transformers

[6] https://huggingface.co/monologg/kobert/tree/main

[7] https://aiopen.etri.re.kr/service_dataset.php

Mehrotra, and Kaheer Suleman. 2017. Frames: a corpus for adding memory to goal-oriented dialogue systems. In *Proceedings of the 18th Annual SIGdial Meeting on Discourse and Dialogue*, pages 207–219, Saarbrücken, Germany, Association for Computational Linguistics.

Maurice Gross. 1997. *The Construction of local grammars.* Finite-State language processing, Roche&Schabes (eds.), MIT Press.

Maurice Gross. 1999. A Bootstrap Method for Constructing Local Grammars. In *Proceedings of the Symposium on Contemporary Mathematics*, University of Belgrade, pages 229–250.

Charles T. Hemphill, John J. Godfrey, and George R. Doddington. (1990). The ATIS Spoken Language Systems Pilot Corpus. In *Proceedings of the workshop on Speech and Natural Language (HLT '90),* pages 96–101, USA. Association for Computational Linguistics,

Changhoe Hwang, Gwanghoon Yoo, and Jeesun Nam. 2021. Construction of Language Resources for Augmenting Intent-annotated Datasets Required for Training Chatbot NLU Models. *Journal of the Language and Information Society*, 44:89–125.

Myeongha Hwang, Jikang Shin, Hojin Seo, Jeong-Seon Im, Hee Cho, and Muhammad Bilal. 2021. KoRASA: Pipeline Optimization for Open-Source Korean Natural Language Understanding Framework Based on Deep Learning. *Journal of the Mobile Information Systems, 1-9.*

Eun-jin Jung. 2005. *Grammaire des adverbes de durée et de date en coréen.* PhD, Université de Marne-la-Vallée.

Xingkun Liu, Arash Eshghi, Pawel Swietojanski, and Verena Rieser. 2019. *Benchmarking natural language understanding services for building conversational agents*. CoRR, abs/1903.05566.

Jeesun Nam. 2018. *An Introduction to a Methodology of Implementing Korean Electronic Dictionaries for Corpus Analysis.* Yeokrak, Seoul

National Institute of Korean Language. *2005. Korean Grammar for Foreigners*. Communication Books. Seoul.

Sébastien Paumier. 2003. *De la reconnaissance de formes linguistiques à l'analyse syntaxique*. Ph.D. theses, Université Paris-Est Marne-la-Vallée, France.

Umutcan Şimşek and Dieter Fensel. 2018. *Intent Generation for Goal-Oriented Dialogue Systems based on Schema.org Annotations*. ArXiv, abs/1807.01292.

Hayssam N. Traboulsi. 2006. *Named Entity Recognition: A Local Grammar-Based Approach*. PhD, University of Surrey.